\documentclass{article}
\usepackage[utf8]{inputenc}
\usepackage{spconf}
\usepackage{color}
\usepackage{bm}

\usepackage{color}
\usepackage{theorem}
\usepackage{times,amsmath,epsfig}
\usepackage{amssymb}
\usepackage{subfigure}
\usepackage{cite}
\usepackage{url}

\usepackage{algorithm, algpseudocode}

\usepackage{tikz}
\usetikzlibrary{shapes,arrows}
\input{mysymbol.sty}

\title{Matrix Low-Rank Trust Region Policy Optimization}
\name{\vspace{-.25cm}Sergio Rozada and Antonio G. Marques\thanks{ Work partially financed by the Spanish NSF Grant PID2019-105032GB-I00 (MCIN/AEI/10.13039/501100011033), by grant TED2021-130347B-I00 (MCIN/AEI/ 10.13039/501100011033 and “European Union NextGenerationEU/PRTR”), by the Autonomous Community of Madrid (CAM - ELLIS Madrid Unit), and by the Young Researchers R\&D Project, ref. num. F861 AUTO-BA-GRAPH (CAM and URJC).}}
\address{Dept. of Signal Theory and Communications, King Juan Carlos University, Madrid, Spain}

\begin{document}
\ninept
\maketitle
\begin{abstract}
Most methods in reinforcement learning use a Policy Gradient (PG) approach to learn a parametric stochastic policy that maps states to actions. The standard approach is to implement such a mapping via a neural network (NN) whose parameters are optimized using stochastic gradient descent. However, PG methods are prone to large policy updates that can render learning inefficient. Trust region algorithms, like Trust Region Policy Optimization (TRPO), constrain the policy update step, ensuring monotonic improvements. This paper introduces low-rank matrix-based models as an efficient alternative for estimating the parameters of TRPO algorithms. By gathering the stochastic policy's parameters into a matrix and applying matrix-completion techniques, we promote and enforce low rank. Our numerical studies demonstrate that low-rank matrix-based policy models effectively reduce both computational and sample complexities compared to NN models, while maintaining comparable aggregated rewards.
\end{abstract}
\begin{keywords}
Reinforcement learning, policy gradients, TRPO, matrix factorization.
\end{keywords}

\section{Introduction}\label{S:Introduction}

In the era of big data, complex dynamical systems call for intelligent algorithms capable of adapting their behavior based on observations. Reinforcement Learning (RL) addresses this issue by learning how to interact with the the world (environment) via trial and error \cite{sutton2018reinforcement, bertsekas2019reinforcement}. More precisely, RL deals with decision-making setups, where an agent must take actions in an environment that changes over time. Traditionally, RL estimates the expected long-term rewards associated with state-action pairs, which are referred to as value function (VF). Then, state-to-action mappings, or policies, are inferred by maximizing the VFs. However, these methods, known as value-based RL, struggle with algorithmic challenges and the curse of dimensionality. 

Alternatively, policy-based methods directly estimate  parametric policies \cite{sutton1999policy}. Policy-based RL maximizes the expected VFs over a close set of parametric policies, normally via \textit{stochastic} gradient ascent schemes. Interestingly, constraining the policy updates to be small guarantees monotonic improvements \cite{kakade2001natural}. This is the basis of Trust Region Policy Optimization (TRPO) \cite{schulman2015trust}. TRPO  iteratively updates the policy within a defined trust region, ensuring minimal policy deviation. One of the main advantages of policy-based methods over value-based ones is that the policies can be probabilistic. States can be mapped into a probability distribution under which actions are drawn from. When dealing with continuous action spaces, Gaussian distributions are commonly used as policy models, with the dependence of the mean and standard deviation with respect of the state being typically modeled using neural networks (NNs) \cite{arulkumaran2017deep}. However, these NN-based policies  frequently encounter convergence difficulties, largely attributable to their reliance on the specific NN architecture. In this paper, we propose an alternative approach by designing a trust-region low-rank (non-NN) method for estimating stochastic policy models, leveraging matrix completion results \cite{eckart1936approximation, markovsky2012low}. Our aim is to design an estimation scheme that is both generic enough to learn the policy and capable of addressing some of the challenges present in NN-based schemes. The rest of this paper is organized as follows. Section 2 introduces the notation commonly used in RL, and frames the TRPO problem to be addressed. Section 3 describes our proposed low-rank TRPO RL approach. Section 4 shows the empirical performance of our algorithm in some classic continuous action space problems, and Section 5 provides concluding remarks.

\vspace{1mm}

\noindent \textbf{Related work and contribution.} Low-rank optimization has been successfully adopted in multiple applications, including matrix completion \cite{eckart1936approximation,markovsky2012low,udell2016generalized,mardani2013decentralized}. While the use of low-rank approaches in RL is less abundant (see, e.g., sparsity-based methods \cite{tolstaya2018nonparametric, lever2016compressed, baek2020sparse}, or linear VF approximation \cite{melo2007q, behzadian2019fast}), there is a growing interest in the topic. Some recent works have focused on leveraging low rank in different elements of the Markov Decision Processes (MDP), such as in the transition probability matrix, or in the reward functions \cite{barreto2016incremental, jiang2017contextual, mahajan2021tesseract}. Also, low rank has been used in the context of VF estimation, initially to compress already estimated VFs \cite{ong2015value}, then to design NN-based estimation schemes \cite{cheng2018factorized, cheng2020novel}. More recently, some works have studied how to leverage factorization techniques to design estimators of the VFs via low-rank matrix \cite{cheng2016co, cheng2017low, shah2020sample, rozada2021low, sam2023overcoming}, and tensor models \cite{rozada2022tensor}. In the context of policy-based methods, however, low-rank has not been extensively studied, although some efforts exist in simple actor-critic setups \cite{rozada2023matrix}. To fill this gap, this paper introduces a \emph{low-rank} design for trust-region policy-based methods via matrix completion. More precisely, i) we focus on TRPO, a trust-region policy-based setup, where we need to estimate the parameters of the policy (actor), and the VFs (critic); ii) we model both, the actor parameters and the critic VF, as matrices; and iii) we enforce low-rank via (tall-times-fat) matrix factorization to regularize the estimation problem. 

\section{Preliminaries}\label{S:Preliminaries}

RL frames the world as a closed-loop setup where agents interact sequentially with a time-indexed environment, defined by a state space $\ccalS$, an action space $\ccalA$, and a reward associated with each state-action pair \cite{sutton2018reinforcement}. Let $t=1,...,T$ be the time index. Given the state $s_t$, the agent takes an action $a_t$, and obtains a numerical signal, or reward $r_t$, quantifying the value of the state-action pair. The goal of RL is to maximize the aggregated (long-term) reward, or return. This poses an interesting problem, as the optimization is coupled across time-steps. The action $a_t$ not only affects the instantaneous reward $r_{t}$, but 
the subsequent states $s_{t'}$ for $t'>t$ and, as a result, future rewards $r_{t'}$ for $t'>t$.  Furthermore, there is a stochastic (non-deterministic) dependence of the reward $r_t$ on the state $s_t$ and action $a_t$.

\noindent \textbf{The actor.} In this context, policy-based RL seeks to learn a parametrized policy $\pi_{\bbtheta} : S \rightarrow A$ that maps states to actions. The policy $\pi_{\bbtheta}$ (strictly speaking, the parameters $\bbtheta$) is estimated by maximizing the expectation of the reward function, that can take different forms \cite{schulman2015high}. A common choice is the state-action value function $Q^{\pi_{\bbtheta}}(s_t, a_t)= \mathbb{E}_{\pi_{\bbtheta}} \left[ G_t | s_t, a_t \right]$, where $G_t=\sum_{t'=t}^T r_{t'}$ is the cumulative reward, or return, and $T$ is a time horizon. However, following this approach leads to high-variance challenges, especially in long-termed setups, i.e. when $T$ is large. A usual strategy to alleviate this problem is to substract a baseline from $Q^{\pi_{\bbtheta}}(s_t, a_t)$, commonly the state-value function $V^{\pi_{\bbtheta}}(s_t)= \mathbb{E}_{\pi_{\bbtheta}} \left[ G_t | s_t \right]$. This residual is defined as the advantage function $A^{\pi_{\bbtheta}}(s_t, a_t) = Q^{\pi_{\bbtheta}}(s_t, a_t) - V^{\pi_{\bbtheta}}(s_t)$, which assesses the expected extra return obtained from taking action $a_t$ over the expected return in state $s_t$ \cite{schulman2015high}. More formally, maximizing the advantage function $A^{\pi_{\bbtheta}}$ leads to the following optimization problem:

\begin{equation}
    \label{eq::pg_maximization}
    \underset{{\bbtheta}}{\text{maximize}} \;\; \mathbb{E}_{\pi_{\bbtheta}} \left[ A^{\pi_{\bbtheta}} (s, a) \right],
\end{equation}

\noindent where the expectation is taken over all state-action pairs. As we don't normally have access to the transition probabilities between states, this problem calls for methods to approximate the expectation $\mathbb{E}_{\pi_{\bbtheta}} \left[ \cdot \right]$. This is usually done by sampling transitions from the environment according to the estimated policy $\pi_{\bbtheta}$. Then, the parameters $\bbtheta$ are updated via \textit{stochastic} gradient ascent \cite{sutton1999policy}.

Alternatively, constraining the policy updates within an arbitrarily small trust-region leads to monotonic policy improvements \cite{schulman2015trust}. This involves recasting the problem in \eqref{eq::pg_maximization} as finding a new policy $\pi_{\bbtheta}$ that maximizes the expected advantage $A^{\pi_{\bbtheta}}$ in a neighbourhood of the actual (old) policy $\pi_{\bbtheta_\text{old}}$ used to sample from the environment. TRPO proposes using the KL-divergence $D_{KL}(\cdot \| \cdot)$ to specify this neighbourhood. More precisely, TRPO constrains the expected KL-divergence between the policy we are optimizing over $\pi_{\bbtheta}$ and the actual policy $\pi_{\bbtheta_\text{old}}$ not to exceed a predefined threshold $\delta$. Formally, let $\bbtheta$ represent the optimization (policy) parameters, and $\bbtheta_\text{old}$ the parameters of the actual policy used to sample from the environment. TRPO proposes to iteratively solve the following maximization: 
\begin{equation}
\label{eq::trpo}
\begin{aligned}
& \underset{{\bbtheta}}{\text{maximize}} &&  \ccalL({\bbtheta}) = \mathbb{E}_{\pi_{\bbtheta_\text{old}}}\left[\frac{\pi_{{\bbtheta}}(a|s)}{\pi_{{\bbtheta}_{\text{old}}}(a|s)}A^{\pi_{{\bbtheta}_{\text{old}}}}(s, a)\right] \\
& \text{subject to} && \mathbb{E}\left[D_{KL}(\pi_{{\bbtheta}_{\text{old}}}(\cdot|s) || \pi_{\bbtheta}(\cdot|s))\right] \le \delta,
\end{aligned}
\end{equation}
\noindent where  $\mathbb{E}_{\pi_{\bbtheta_\text{old}}}\left[ \cdot \right]$ denotes the expectation over the states, and the actions, and $\mathbb{E}\left[ \cdot \right]$ denotes the expectation only over the states. 

Considering that the transition probabilities are unknown, TRPO involves two iterative steps: i) sample the environment following the policy $\pi_{\bbtheta_\text{old}}$ to approximate the expectation $\mathbb{E}_{\pi_{\bbtheta_\text{old}}}\left[ \cdot \right]$, and ii) use the samples in i) to solve an approximated version of \eqref{eq::trpo}. Moreover, since solving the sample-based approximated version of \eqref{eq::trpo} is non-trivial, a common approach is further simplify the problem using Taylor expansions \cite{kakade2001natural}. In particular, using a first-order approximation for the objective and a second-order approximation for the constraint leads to the following quadratic problem:

\begin{equation}
\label{eq::trpo_relaxed}
\begin{aligned}
    & \underset{\bbtheta}{\text{maximize}} && \bbg^T (\bbtheta - \bbtheta_\text{old}) \\
    & \text{subject to} && (\bbtheta - \bbtheta_\text{old})^T \bbH (\bbtheta - \bbtheta_\text{old}) \le \delta,
\end{aligned}
\end{equation}
where $\bbg := \nabla_{\bbtheta} \ccalL({\bbtheta}) \left. \right|_{\bbtheta = \bbtheta_\text{old}}$ is the gradient of the cost function w.r.t. the set of parameters $\bbtheta$ evaluated at $\bbtheta_\text{old}$, and $\bbH$ is the Hessian matrix of the KL constrain evaluated at $\bbtheta_\text{old}$. More precisely, $\bbH_{i, j} = \frac{\partial }{\partial {\bbtheta}_i} \frac{\partial } {\partial {\bbtheta}_j} \mathbb{E}_{t}\left[D_{KL}(\pi_{{\bbtheta}_{\text{old}}}(a|s) || \pi_{\bbtheta}(a|s))\right]\left. \right|_{\bbtheta = \bbtheta_\text{old}}$. Interestingly, the gradient of the constraint evaluated at ${\bbtheta}={\bbtheta}_\text{old}$ is zero and the Hessian evaluated at ${\bbtheta}={\bbtheta}_\text{old}$ is the Fisher information matrix (FIM) \cite{amari2012differential}. Thus, we can redefine $\bbH$ in terms of the policy scores $\nabla_{\bbtheta} \text{log} \; \pi_{{\bbtheta}}$ as $\bbH = \mathbb{E}_{\pi_{\bbtheta_\text{old}}}\left[ \nabla_{\bbtheta} \text{log} \; \pi_{{\bbtheta}}(a|s) \nabla_{\bbtheta} \text{log} \; \pi_{{\bbtheta}}(a|s)^T \right] \left. \right|_{\bbtheta = \bbtheta_\text{old}}$ as per the FIM definition. Trivially, the gradient  $\bbg$ can also be reformulated in terms of the policy scores $\nabla_{\bbtheta} \text{log} \; \pi_{{\bbtheta}}$ as $\bbg = \mathbb{E}_{\pi_{\bbtheta_\text{old}}}\left[\frac{\pi_{{\bbtheta}}(a|s)}{\pi_{{\bbtheta}_{\text{old}}}(a|s)} \nabla_{\bbtheta} \text{log} \; \pi_{{\bbtheta}}(a|s) A^{\pi_{{\bbtheta}_{\text{old}}}}(s, a)\right]\Big|_{\bbtheta = \bbtheta_\text{old}}$. Lastly, as we use samples to approximate the expectations, we need to form sto- chastic estimates of the gradient $\bbg$, and the Hessian $\bbH$, leading to:
\begin{align}
    \label{eq::pg_gradient}
    \bbg &\approx \mathbb{E}_{t}\left[\frac{\pi_{{\bbtheta}}(a_t|s_t)}{\pi_{{\bbtheta}_{\text{old}}}(a_t|s_t)} \nabla_{\bbtheta} \text{log} \; \pi_{{\bbtheta}}(a|s) A^{\pi_{{\bbtheta}_{\text{old}}}}(s_t, a_t)\right]\Bigg|_{\bbtheta = \bbtheta_\text{old}}, \\
    \label{eq::pg_hessian}
    \bbH &\approx \mathbb{E}_{t}\left[ \nabla_{\bbtheta} \text{log} \;\pi_{{\bbtheta}}(a_t|s_t) \nabla_{\bbtheta} \text{log} \;\pi_{{\bbtheta}}(a_t|s_t)^T \right] \Big|_{\bbtheta = \bbtheta_\text{old}}.
\end{align}

\noindent \textbf{The critic.} The advantages $A^{\pi_{{\bbtheta}_{\text{old}}}}$ are needed to form the stochastic gradients in \eqref{eq::pg_gradient}. However, its exact estimation is a key problem in value-based RL, and it is well-known that it suffers from the curse of dimensionality. To overcome this issue, one common approach is to use a stochastic parametric estimate of the advantage $A_\omega(s_t) = G_t - V_\omega(s_t)$, where $G_t$ is a sample return, and $V_\omega$ is a parametric model of the VF. This leads to a two-step actor-critic method that alternates the estimation of the parameters of the policy $\pi_{\bbtheta}$ (actor) as well as those of the VF $V_\omega$ (critic). Given a collection of samples, the parameters of the actor $\bbtheta$ are updated by solving the maximization problem defined in \eqref{eq::trpo_relaxed}. The gradients $\bbg$, and the Hessian $\bbH$ are formed using the stochastic gradients defined in \eqref{eq::pg_gradient}, where the advantage function is approximated as $A^{\pi_{{\bbtheta}_{\text{old}}}} (s_t, a_t) \approx G_t - V_\omega(s_t)$, and $V_\omega(s_t)$ is considered given. On the other hand, the parameters $\bbomega$ of the VF are obtained from solving
\begin{equation}
    \label{eq::critic_loss}
    {\bbomega}^*=\arg\min_{\bbomega} \ccalL(\bbomega) := \arg\min_{\bbomega} \frac{1}{2}\sum_{t=0}^T (G_t - V_{\bbomega}(s_t))^2,
\end{equation}
\noindent typically using gradient descent methods of the form 
\begin{equation}
\bbomega_{h+1}=\bbomega_{h} - \alpha_\omega \nabla_{\bbomega} \ccalL(\bbomega_h),
\end{equation} 
where $h$ is the iteration index and $\alpha_\omega$ is the gradient step.

\section{Low-Rank TRPO}\label{S:Contribution}

According to the previous discussion, TRPO boils down to postulating the adequate underlying model of the policy $\pi_{\bbtheta}$, and obtaining the policy scores $\nabla_{\bbtheta} \text{log} \; \pi_{{\bbtheta}}$. As mentioned earlier, adoption of univariate Gaussian policies $a_t\sim \ccalN(a | \mu(s_t), \sigma(s_t))$ is common when $\ccalA$ is continuous \cite{ciosek2018expected} and, as a result, the goal becomes finding the mean $\mu(s_t)$ and the standard deviation $\sigma(s_t)$ associated with each state\footnote{Alternatively, many RL works set the standard deviation to be the same across states and focus on learning $\mu(s_t)$.} $s_t$. These functions $\mu:\ccalS \mapsto \reals$ and $\sigma:\ccalS \mapsto \reals^+$ are usually parametric, and oftentimes implemented by NNs \cite{arulkumaran2017deep}. In this work, however, we propose a low-rank non-parametric approach for designing the mean $\mu(s_t)$ and the standard deviation $\sigma(s_t)$ mappings. More specifically, we first codify every state $s \in \ccalS$ using a tuple of indices $(i_s,j_s)$ (suppose for simplicity that we deal with an RL scenario where the state space has two dimensions) and, then, use matrices to represent mappings from $\ccalS \mapsto \reals$. More precisely, for the case of Gaussian policies, we define two matrices $\bbX_\mu\in\reals^{N\times M}$ and $\bbX_\sigma\in\reals^{N\times M}$, which collect, for all $NM$ states, the associated means and standard deviations. Under this approach, when $s$ is observed, the associated action is drawn from a Gaussian with mean $[\bbX_\mu]_{i_s,j_s}$ and standard deviation $[\bbX_\sigma]_{i_s,j_s}$. We leverage low-rank in $\bbX_\mu$ and $\bbX_\sigma$ via matrix factorization \cite{markovsky2012low}. In particular, we 1) introduce matrices $\bbL_\mu$ and $\bbL_\sigma$ (tall), and $\bbR_\mu$ and $\bbR_\sigma$ (fat); and 2) rewrite the original matrices as $\bbX_\mu=\bbL_\mu \bbR_\mu$ and $\bbX_\sigma=\bbL_\sigma \bbR_\sigma$. The benefit of this approach is two-fold, we reduce the number of parameters to estimate, and we ease the estimation from a limited number of observations. More explicitly, the mappings from the state to the Gaussian parameters under the proposed low-rank models are:
\vspace{-.5mm}

\begin{equation}
\label{eq:mu_sigma_polynomial_order_two}
\begin{aligned}
 \mu (s_t)\! &= \!\textstyle \sum_{k=1}^K [\bbL_\mu]_{i{s_t}, k} [\bbR_\mu]_{k, j{s_t}} \\ \sigma (s_t)\! &=\! \textstyle \sum_{k=1}^K [\bbL_\sigma]_{i{s_t}, k} [\bbR_\sigma]_{k, j{s_t}}.&
\end{aligned}
\end{equation}

In our approach, the parameters $\bbtheta$ are the entries of the matrices $\{\bbL_\mu, \bbR_\mu, \bbL_\sigma, \bbR_\sigma\}$. Our goal is to find an expression for the entries of the policy score $\nabla_{\bbtheta}  \text{log} \; \pi_{\bbtheta}$, where the underlying policy is Gaussian. Consider first $\bbL_\mu$ and $\bbR_\mu$. We look for an expression for the entries of $\nabla_{\bbL_\mu}  \text{log} \; \ccalN(a_t | \mu(s_t), \sigma(s_t))$, and  $\nabla_{\bbR_\mu}  \text{log} \; \ccalN(a_t | \mu(s_t), \sigma(s_t))$. We apply the chain rule, combining the partial derivatives of the Gaussian probability distribution w.r.t. the mean function $\mu$, with the partial derivatives of the $\mu$ function in   \eqref{eq:mu_sigma_polynomial_order_two} w.r.t. the entries of the matrices $\bbL_\mu$, and $\bbR_\mu$:
\begin{align}
    \frac{\partial \; \text{log} \; \ccalN(a_t | \mu(s_t), \sigma(s_t))}{\partial [\bbL_\mu]_{i, k}} =  
     \mathbb{I}_{i=i_{s_t}}  \nonumber \\ \frac{a_t - [\bbL_\mu \bbR_\mu]_{i_{s_t}, j_{s_t}}}{[\bbL_\sigma \bbR_\sigma]_{i_{s_t}, j_{s_t}}^2} [\bbR_\mu]_{k, j_{s_t}} \label{eq::pg_l_mu} \\
    \frac{\partial \; \text{log} \; \ccalN(a_t | \mu(s_t), \sigma(s_t))}{\partial [\bbR_\mu]_{k, j}} = 
     \mathbb{I}_{j=j_{s_t}}  \nonumber \\
    \frac{a_t - [\bbL_\mu \bbR_\mu]_{i_{s_t}, j_{s_t}}}{[\bbL_\sigma \bbR_\sigma]_{i_{s_t}, j_{s_t}}^2} [\bbL_\mu]_{i_{s_t}, k} \label{eq::pg_r_mu} 
\end{align}
\noindent where $\mathbb{I}_{i=i_{s_t}}$, and $\mathbb{I}_{j=j_{s_t}}$ are indicator functions. Similarly, we need the derivatives of the Gaussian probability distribution w.r.t. the $\sigma$ function, 
and the derivatives of  $\sigma$ function in \eqref{eq:mu_sigma_polynomial_order_two} w.r.t. the entries of $\bbL_\sigma$ and $\bbR_\sigma$:
\begin{align}
    \frac{\partial \; \text{log} \; \ccalN(a_t | \mu(s_t), \sigma(s_t))}{\partial [\bbL_\sigma]_{i, k}} =
    \mathbb{I}_{i=i_{s_t}}  \nonumber \\
    \left( \frac{(a_t - [\bbL_\mu \bbR_\mu]_{i_{s_t}, j_{s_t}})^2}{2[\bbL_\sigma \bbR_\sigma]_{i_{s_t}, j_{s_t}}^3} - \frac{1}{[\bbL_\sigma \bbR_\sigma]_{i_{s_t}, j_{s_t}}}\right) [\bbR_\sigma]_{k, j_{s_t}} \label{eq::pg_l_sigma} \\
   \frac{\partial \; \text{log} \; \ccalN(a_t | \mu(s_t), \sigma(s_t))}{\partial [\bbR_\sigma]_{k, j}} =
     \mathbb{I}_{j=j_{s_t}}  \nonumber \\
    \left( \frac{(a_t - [\bbL_\mu \bbR_\mu]_{i_{s_t}, j_{s_t}})^2}{2[\bbL_\sigma \bbR_\sigma]_{i_{s_t}, j_{s_t}}^3} - \frac{1}{[\bbL_\sigma \bbR_\sigma]_{i_{s_t}, j_{s_t}}}\right) [\bbL_\sigma]_{i_{s_t}, k} \label{eq::pg_r_sigma} 
\end{align}

Finally, we need to estimate the parameters of the critic $\omega$ too. Again, we collect the VFs in a matrix $\bbX_\omega \!\in\! \reals^{N\times M}$ to then postulate that the VF matrix is low rank, and factorized as the product of $\bbL_\omega \!\in \! \reals^{N\times K}$ and $\bbR_\omega \!\in\! \reals^{K\times M}$ with $K\ll$ $\min\{N,M\}$. Consequently, the critic takes the polynomial form $V(s_t) \!= \!\!\sum_{k=1}^K [\bbL_\omega]_{i_{s_t}\!, k} [\bbR_\omega]_{k, j_{s_t}\!}$. As stated previously, the parameters of the critic $\omega$ are found via gradient descent, using the partial derivatives of the critic cost in  \eqref{eq::critic_loss} w.r.t. the entries of ${\bbL_\omega}$ and  $\bbR_\omega$
\begin{align}
    \frac{\partial \ccalL(\bbL_\omega, \bbR_\omega)}{\partial [\bbL_\omega]_{i, k}} = \sum_{t=0}^T \mathbb{I}_{i=i_{s_t}} (G_t - [\bbL_\omega \bbR_\omega]_{i_{s_t}, j_{s_t}})[\bbR_\omega]_{k, j_{s_t}} \label{eq::pg_l_omega}
\end{align}
\begin{align}
    \frac{\partial \ccalL(\bbL_\omega, \bbR_\omega)}{\partial [\bbR_\omega]_{k, j}} = \sum_{t=0}^T \mathbb{I}_{j=j_{s_t}} (G_t - [\bbL_\omega \bbR_\omega]_{i_{s_t}, j_{s_t}})[\bbL_\omega]_{i_{s_t}, k} \label{eq::pg_r_omega}
\end{align}

\noindent \textbf{The algorithm.} Now, we can formulate a trust-region low-rank policy optimization (TRLRPO) algorithm. The agent samples transitions from the environment using the Gaussian policy $\pi_{\mu, \sigma}=\ccalN(a_t | \mu(s_t), \sigma(s_t))$, where the mean is $\mu(s_t)=[\bbL_\mu \bbR_\mu]_{i_{s_t}, j_{s_t}}$, and the standard deviation is $\sigma(s_t)=[\bbL_\sigma \bbR_\sigma]_{i_{s_t}, j_{s_t}}$. Then, we build $\bbg$, and $\bbH$. To such extent, we evaluate \eqref{eq::pg_l_mu}--\eqref{eq::pg_r_sigma} with the current estimates of the matrices $\{\bbL_\mu, \bbR_\mu, \bbL_\sigma, \bbR_\sigma\}$, and average across samples. Finally, we solve the problem in \eqref{eq::trpo_relaxed} to obtain our new matrices $\bbL_\mu$, $\bbR_\mu$, $\bbL_\sigma$, and $\bbR_\sigma$. This is normally done via conjugate gradient ascent methods. We complete an iteration by updating the critic matrices $\bbL_\omega$, and $\bbR_\omega$ with a gradient descent step using the derivatives defined in \eqref{eq::pg_l_omega} and \eqref{eq::pg_r_omega}. The algorithm is depicted in Algorithm \eqref{alg:trlrpo}. Low-rank policies are very efficient in terms of parameters in comparison with NNs, enhancing the convergence speed. When the state space is smooth (i.e., the states are similar) and high-dimensional, low-rank policies can help overcoming the stability problems of NNs. 

\begin{algorithm}[!htbp]
\flushleft
\caption{Trust-Region Low-Rank Policy Optimization}\label{alg:trlrpo}
\begin{algorithmic}
    \Require Initial policy and VF matrices $\bbL_\mu^0, \bbR_\mu^0, \bbL_\sigma^0$, $\bbR_\sigma^0$,  $\bbL_\omega^0$, and $\bbR_\omega^0$; critic learning rate $\alpha_\omega$; maximum number of iterations $H$; and maximum number of episodes per iteration $E$.
    \For{$h=0, ...,  H$}
        \State{Initialize empty buffer $B$ to store samples}
        \For{$e=0, ...,  E$} \hfill\Comment{Sample transitions}
            \State{Observe initial state $s_0$}
            \For{$t=0, ..., T$} 
                \State{$\mu_{s_t} \gets [\bbL_\mu \bbR_\mu]_{i_{s_t}, j_{s_t}}$}
                \State{$\sigma_{s_t} \gets [\bbL_\sigma \bbR_\sigma]_{i_{s_t}, j_{s_t}}$}
                \State{$a_t \sim \ccalN(a|\mu_{s_t}, \sigma_{s_t})$}
                \State{Take action $a_t$, observe next state $s_{t'}$, and reward $r_t$}
                \State{Append the tuple $(s_t, a_t, s_{t'}, r_t)$ to the buffer $B$}
                \State{$s_t \gets s_{t'}$}
            \EndFor
        \EndFor
        \State
        \State Form $\bbg^h$ using the derivatives in \eqref{eq::pg_l_mu}--\eqref{eq::pg_r_sigma} \hfill\Comment{Actor update}
        
        \State Form $\bbH^h$ using the derivatives in \eqref{eq::pg_l_mu}--\eqref{eq::pg_r_sigma} 

        \State $\bbL_\mu^{h+1}, \bbR_\mu^{h+1},\bbL_\sigma^{h+1}, \bbR_\sigma^{h+1} \gets$ Solve \eqref{eq::trpo_relaxed}
        
        \State
        \State $\bbL_\omega^{h+1} \gets \bbL_\omega^h + \alpha_\omega \nabla_{\bbL_\omega} J(\bbL_\omega^h, \bbR_\omega^h)$ \hfill\Comment{Critic update}

        \State $\bbR_\omega^{h+1} \gets \bbR_\omega^h + \alpha_\omega \nabla_{\bbR_\omega} J(\bbL_\omega^h, \bbR_\omega^h)$ 
    \EndFor
\end{algorithmic}
\end{algorithm}

\vspace{.2cm}
\noindent\textbf{Remark:} To simplify exposition and obey page constraints, we have focused on how to apply our approach to RL setups with continuous actions and Gaussian policies. However, the low-rank TRPO approach described in this section can also be extended to other policy models, including those dealing with discrete actions. In such a case, a probabilistic softmax policy is implemented. Succinctly, this alternative design would imply i) collecting a matrix of weights $\bbX_\bbz \in \reals^{|\ccalS|\times |\ccalA|}$, where the i-th row is a vector $\bbz_i$ containing the softmax scores of a given state; ii) proposing a low-rank (tall-times-fat) matrix model of the form $\bbX_\bbz = \bbL_\bbz \bbR_\bbz$; and iii) deriving the corresponding policy scores for the softmax function.

\section{Numerical experiments}\label{S:Simulations}

We have tested our TRLRPO algorithm in three continuous-action problems of the toolkit OpenAI Gym \cite{brockman2016openai}: i) the inverted pendulum, where an agent tries to keep a pendulum upright; ii) the acrobot, a double pendulum that needs to swing up; and iii) the mountain car, which tries to get impulse to reach the top of a hill. TRLRPO is bench-marked against TRPO with NN-based policy models (NN-TRPO). We have compared the i) efficiency in terms of parameters, ii) the convergence rate, and iii) the return obtained by the estimated policies. The exact implementation details, together with additional test cases and experiments, can be found in \cite{rozada2023code}.

\vspace{1mm}

\noindent \textbf{Experimental setup.} As the action spaces of all setups are continuous, the action $a_t$ in the state $s_t$ is sampled from the normal distribution $\ccalN(a|\mu(s_t), \sigma)$. The estimation of the mean $\mu(s_t)$ is the main difference between TRLRPO and NN-TRPO. As customary in many PG setups, we have eliminated the dependence of $\sigma$ on $s_t$ for the sake of simplicity. State spaces are normally continuous, and, while NNs can deal with them, searching for the proper architecture can be challenging. TRLRPO, on the other hand, discretizes the state space $\ccalS$. Discretization trades-off resolution and performance. The finer the discretization, the larger the number of entries of $\bbX_\mu$ and $\bbX_\omega$. The key aspect of low rank is that it can keep fine sampling resolutions while reducing the number of parameters. In this setup, we have sampled a regular grid over the state space $\ccalS$. The size of the state space is then  defined by the Cartesian product of the sampled grid. To estimate the advantage function $A^{\pi_{\bbtheta}}$, both algorithms implement a critic step to learn the VFs. To be consistent with the proposed approach, TRLRPO models the VF as $V(s_t)=[\bbL_\omega \bbR_\omega]_{i_{s_t}, j_{s_t}}$, while NN-TRPO models the VFs as $V_\omega(s_t)=\text{NN}_\omega(s_t)$. We run $100$ simulations in each scenario, and we measure the return per episode $\check{\ccalR}=\sum_{t=0}^T r_t$. To compare more fairly the number of parameters required by both approaches, in all setups we tested several fully-connected NNs (each with a different size) and reported the results of the smallest one that solved the problem.

\vspace{1mm}

\begin{figure}[h]
    \centering
    \includegraphics[width=.9\linewidth]{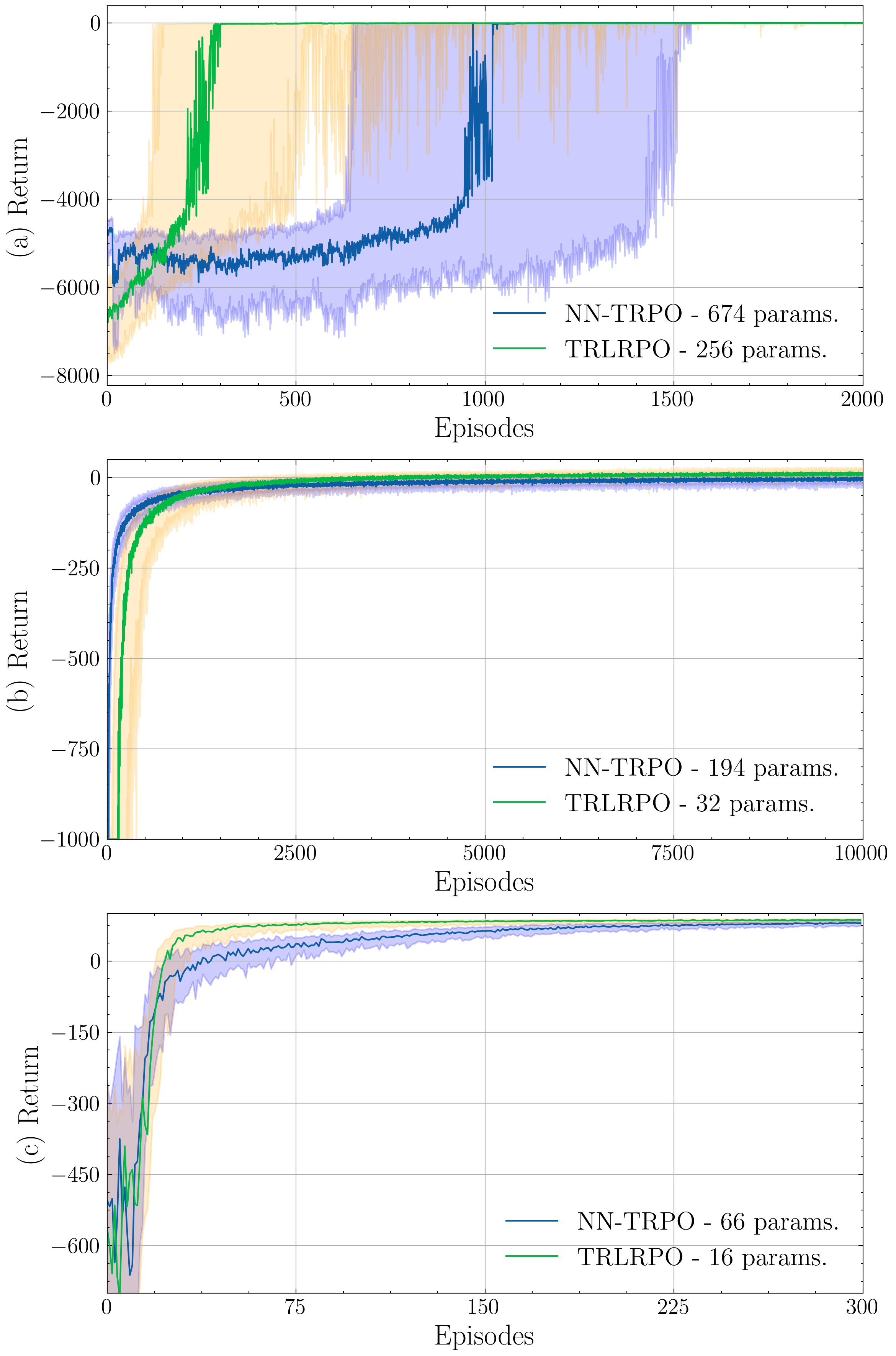}
    
    \vspace{-3mm}
    \caption{Median return per episode in 3 standard RL problems: (a) the pendulum, (b) the acrobot, and (c) the mountain car. The number of parameters of each model is shown in the legend. TRLRPO reaches the steady state faster than NN-TRPO in the pendulum, and mountain car problems, achieving a better return in the acrobot problem.}
    \label{fig::cumreward}
    \vspace{-5mm}
\end{figure}

\noindent \textbf{Analysis of results.} Fig. \ref{fig::cumreward} presents the median $\check{\ccalR}$ across the $100$ simulations. The first observation is that in the pendulum and mountain car problems, TRLRPO achieves a steady state faster than NN-TRPO. More precisely, TRLRPO reaches its maximum return around episode $250$ in the pendulum problem, while NN-TRPO stabilizes around episode $1,100$. In the mountain car problem, TRLRPO reaches a steady state around episode $100$, and NN-TRPO around episode $300$. Although NN-TRPO initially learns faster in the acrobot problem, it is not clear which algorithm stabilizes before. However, TRLRPO obtains higher returns. In contrast, in the pendulum and mountain-car problems, the final returns obtained by both algorithms are similar. The simplicity and parametrization-efficiency of low-rank policy models are likely the reason why TRLRPO converges (reaches steady state) faster than NN-TRPO, leading to the same (or better) returns faster than other alternatives.

Regarding the number of parameters, the panels in Fig. \ref{fig::cumreward} show that TRLRPO needs significantly less parameters to reach similar/higher returns than NN-TRPO. As stated in the description of the experimental setup, we recall that the space of fully-connected NN was searched to look for the smallest architecture (in terms of number of parameters) that converges while achieving good results. With this in mind, the experiments show that in the pendulum environment, the size of the TRLRPO model is $~38\%$ the size of the NN-TRPO model ($62\%$ savings). The savings are even larger in the acrobot and mountain-car setups, where the TRLRPO models need approximately $~16\%$, and $~24\%$ of the parameters employed by their NN counterparts. 

\section{Conclusions}
\label{S:conclusions}

This paper presents a trust-region \emph{low-rank} policy optimization (TRLRPO) algorithm that leverages low-rank in the context of policy-based Reinforcement Learning. More precisely, we use matrix completion techniques to postulate Gaussian policy models in continuous action setups. Discretization is a simple yet effective approach to deal with continuous action spaces. However, fine sampling resolutions lead to large parameter spaces. Low rank helps balancing the sampling resolution, and the size of the model. We compared TRLRPO against NN-based TRPO (NN-TRPO) in three OpenAI Gym continuous action environments. TRLRPO and NN-TRPO achieve similar returns, but TRLRPO saves a significant number of parameters. Furthermore, and partly due to the smaller size of the policy models, TRLRPO reaches the steady state faster than NN-TRPO. In summary, this paper shows that low rank is a promising tool to design parsimonious policy-based RL algorithms. Future research directions include the theoretical characterization of low-rank policy-based methods, and generalizations to highly dimensional environments via low-rank tensor decomposition.

\bibliographystyle{IEEEbib}
\bibliography{myIEEEabrv, biblio}

\end{document}